\documentclass[journal,twoside,web]{ieeecolor}
\usepackage{etoolbox}
\makeatletter
\@ifundefined{color@begingroup}%
{\let\color@begingroup\relax
	\let\color@endgroup\relax}{}%
\def\fix@ieeecolor@hbox#1{%
	\hbox{\color@begingroup#1\color@endgroup}}
\patchcmd\@makecaption{\hbox}{\fix@ieeecolor@hbox}{}{\FAILED}
\patchcmd\@makecaption{\hbox}{\fix@ieeecolor@hbox}{}{\FAILED}

\usepackage{tmi}
\usepackage{cite}
\usepackage{amsmath,amssymb,amsfonts}
\usepackage{algorithmic}
\usepackage{graphicx}
\usepackage{textcomp}
\usepackage{booktabs}
\usepackage{multirow}
\makeatletter
\let\NAT@parse\undefined
\makeatother
\usepackage[colorlinks,
linkcolor=black,
anchorcolor=black,
citecolor=black]{hyperref}

\def\journalname{xxxxxx}

\def\BibTeX{{\rm B\kern-.05em{\sc i\kern-.025em b}\kern-.08em
    T\kern-.1667em\lower.7ex\hbox{E}\kern-.125emX}}
\begin{document}
\title{A Heterogeneous Graph Neural Network Fusing Functional and Structural Connectivity for MCI Diagnosis}
\author{Feiyu Yin, Yu Lei, Siyuan Dai, Wenwen Zeng, Guoqing Wu, Liang Zhan, and Jinhua Yu, \IEEEmembership{Member, IEEE}
\thanks{This work was supported in part by the National Natural Science Foundation of China under Grant 82372096, and in part by the National Key Research and Development Program of China under Grant 2023YFC251000. (Feiyu Yin and Yu Lei contributed equally to this work.) (Corresponding authors: Jinhua Yu.)}
\thanks{Feiyu Yin and Yu Lei are with the School of Information Science and Technology, Fudan University, Shanghai, 200433, China. (e-mail:yinfy0822@163.com).}
\thanks{Wenwen Zeng and Guoqing Wu are with the School of Information Science and Technology, Fudan University, Shanghai, 200433, China. }
\thanks{Siyuan Dai and Liang Zhan, Department of Electrical and Computer Engineering, University of Pittsburgh, Pittsburgh, 15213, USA.} 
\thanks{Jinhua Yu is with the School of Information Science and Technology, Fudan University, Shanghai, 200433, China. (e-mail: jhyu@fudan.edu.cn).}}

\maketitle

% 去掉页眉  
\thispagestyle{empty} % 使当前页没有页眉

\begin{abstract}
Brain connectivity alternations associated with brain disorders have been widely reported in resting-state functional imaging (rs-fMRI) and diffusion tensor imaging (DTI). While many dual-modal fusion methods based on graph neural networks (GNNs) have been proposed, they generally follow homogenous fusion ways ignoring rich heterogeneity of dual-modal information. To address this issue, we propose a novel method that integrates functional and structural connectivity based on heterogeneous graph neural networks (HGNNs) to better leverage the rich heterogeneity in dual-modal images. We firstly use blood oxygen level dependency and whiter matter structure information provided by rs-fMRI and DTI to establish homo-meta-path, capturing node relationships within the same modality. At the same time, we propose to establish hetero-meta-path based on structure-function coupling and brain community searching to capture relations among cross-modal nodes. Secondly, we further introduce a heterogeneous graph pooling strategy that automatically balances homo- and hetero-meta-path, effectively leveraging heterogeneous information and preventing feature confusion after pooling. Thirdly, based on the flexibility of heterogeneous graphs, we propose a heterogeneous graph data augmentation approach that can conveniently address the sample imbalance issue commonly seen in clinical diagnosis. We evaluate our method on ADNI-3 dataset for mild cognitive impairment (MCI) diagnosis. Experimental results indicate the proposed method is effective and superior to other algorithms, with a mean classification accuracy of 93.3\%.
\end{abstract}


%\section{\textcolor{black}{Introduction}}
\section{Introduction}
\label{sec:introduction}
\IEEEPARstart{M}{agnetic} Resonance Imaging (MRI) has emerged as a valuable tool in neuroscience, offering deeper objective insights into neurological disorders and their underlying mechanisms. Previous researches have demonstrated the valuable contributions of resting-state functional MRI (rs-fMRI) and Diffusion Tensor Imaging (DTI) among multiple MRI modalities in understanding brain structure and function \cite{b1,b2}. Specifically, brain functional connectivity (FC) constructed from rs-fMRI imaging can capture spontaneous neuronal activity and reveal intrinsic connections between brain regions \cite{b3,b4} while brain structural connectivity (SC) constructed from DTI imaging is able to provide crucial insights into the integrity of white matter structures and the identification of neural fiber abnormalities \cite{b5,b6}. In recent years, many studies find that alterations in neuronal functioning are directly related to alterations in white matter structure \cite{b7,b8}, which has given rise to a lot of work on dual-modal fusion of FC and SC.\par
Since FC and SC can be conveniently described by graphs, framework based on graph neural networks (GNNs) have become a popular choice to identify brain disorders combining dual-modal information \cite{b9,b10,b11,b12}. There are mainly two popular ways to fuse dual-modal information with GNNs, which are feature-level fusion and connectivity-level fusion. Specifically, in feature-level fusion, the same backbone is applied to different modalities to extract functional and structural features separately, then features are concatenated or weighted summed together as the fused feature for further analysis. While in connectivity-level fusion, usually a summative homogeneous graph (i.e., graph that have only one type of node and one type of edge) will be constructed based on fused structural-functional connectivity, and the fused structure-function features will be extracted from the summative graph. While these approaches are able to fuse dual-modal features, there is more information of connectivity yet to be explored. Firstly, there is no feature interaction in the above feature fusion methods, which leads to insufficient feature learning and reduced classification performance. Although connectivity-level fusion provides more consistent feature embedding than feature-level fusion, the summative graphs constructed in homogeneous way may disrupt the inherent heterogeneity between FC and SC, such as differences in feature spaces and graph topologies. These problems urge a new dual-modal fusion approach to better synthesize the information from FC and SC to identity brain disorders.\par
Heterogeneous graph (HG) provides new ways of describing the complex connectivity in real world \cite{b13}, which drives the development of heterogeneous graph neural networks (HGNNs) in fields such as social network analysis and bioinformatics \cite{b13,b14}. Considering that integrating FC and SC into HG can effectively preserve the heterogeneous information in the two modalities, we pursue a new dual-modal fusion method based on HG in present work. There are several challenges needed to be addressed yet. Specifically, i) different types of relations in HG are usually defined through meta-path which is semantic dependent \cite{b15,b16}, meaning that meta-path needs to follow the inherent connectivity within the modality as well as to reveal interactive connectivity between modalities. ii) Pooling strategies for HGs need to cope with more complex relations, and directly applying pooling strategies designed for homogeneous graphs \cite{b17,b18} can lead to feature confusion among different types of nodes in HGs. iii) Differences in the incidence of various brain diseases lead to sample imbalance, which in turn affects the classification performance of graph networks.\par
Therefore, we propose several effective mechanisms to address these challenges. Firstly, in constructing HG, in order to capture node relationships within rs-fMRI or DTI modality and relationships among cross-modal node pairs, we propose to define homo-meta-path and hetero-meta-path. Based on existing rs-fMRI and DTI studies, we naturally utilize blood oxygen level dependency information to construct FC and white matter fibers to construct SC, which serve as homo-meta-paths. On the other hand, we propose to establish hetero-meta-path from node-level and community-level based on structure-functional coupling \cite{b19} and brain community searching \cite{b20} to capture cross-modal relationships. Secondly, as existing fusion methods based on GNNs are not capable for HGs, we introduce a novel HG pooling strategy that not only comprehensively considers heterogeneous topology but also can avoid feature confusion among different types of nodes. Thirdly, to address the common issue of sample imbalance in brain disorder datasets, we propose a novel HG augmentation method leveraging the adaptability of HG.\par
The main contributions of our present work can be summarized as follows:\par
1) We propose a novel HGNN to fuse rs-fMRI and DTI information for MCI diagnosis. In constructing HG, we propose to establish homo-meta-path reflecting unimodal connectivity and hetero-meta-path reflecting dual-modal inter-relations, where structure-function coupling and brain community searching are used in establishing hetero-meta-path.\par
2) We introduce a novel HG pooling strategy which can automatically balance homo- and hetero-meta-path, effectively leveraging heterogeneous information and preventing feature confusion after pooling.\par
3) We propose an HG augmentation method leveraging the adaptability of HG to address the issue of sample imbalance, which is a common factor that affects the performance of diagnostic model in classification of brain disorders.\par
The proposed method is validated using a mild cognitive impairment (MCI) dataset sampled from ADNI-3 dataset. Experimental results indicate that our method can achieve remarkable performance for MCI identification.\par
The structure of this paper is organized as follows. Section \ref{sec:background} introduces the most relevant concepts. In Section \ref{sec:materials}, we introduce materials used in this work. Section \ref{sec:methods} introduces details of proposed method. In Section \ref{sec:experiments and results}, we introduce experimental settings, state-of-the-art methods and present experimental results. Section \ref{sec:discussion} discusses the influence of proposed key mechanisms in our method. We conclude this letter in Section \ref{sec:conclusion}.\par

%\section{\textcolor{black}{Background}}
\section{Background}
\label{sec:background}

\subsection{Heterogeneous Graph}
\subsubsection{Heterogeneous Graph \cite{b21}}
A HG denoted as $G=\left\{V, E\right\}$ consists of a node set $V$ and an edge set $E$, respectively. Considering that different node types and edge types exit in HG, the set $\mathcal{N}$ of node types and a set $\mathcal{E}$ of edge types need to be predefined. Then we can build mapping function between nodes and node types as $\phi: V \rightarrow \mathcal{N}$, and mapping function between edges and edge types as $\psi: E \rightarrow \mathcal{E}$. It’s significant to stress that $|\mathcal{N}|+|\mathcal{E}|>2$.

\subsubsection{Meta-path \cite{b22}}
A meta-path $\Phi$ is defined as a path in a form of $\mathcal{N}_1 \xrightarrow{R_1} \mathcal{N}_2 \xrightarrow{R_2} \cdots \xrightarrow{R_l} \mathcal{N}_{l+1}$, which describes a composite relation $R=R_1{ }^{\circ} R_2{ }^{\circ} \cdots{ }^{\circ} R_l$ between node types $\mathcal{N}_1$ and $\mathcal{N}_{l+1}$. Given a meta-path $\Phi$, there exists a set of meta-path based neighbors which can reveal diverse structure and rich semantic information in HG. Take a node $v_i$ and a meta-path $\Phi$ for example, the meta-path based neighbors $\mathcal{N}_i^\Phi$ of node $v_i$ are defined as the set of nodes which connect with node $v_i$ via meta-path $\Phi$. Note that the node’s neighbors include itself.

\subsection{Heterogeneous Graph Attention Network (HAN)}
HAN \cite{b23} was first proposed in the field of social network analysis based on the hierarchical attention, including node-level and semantic-level attentions to extract the heterogeneity and rich semantic information from heterogeneous graph.\par
For a node $v_i$ in a HG, we denote its node feature of the $l$-th layer as $h_i^{(l)}$, the node-level and semantic-level attentions for meta-path $\Phi_p$ as $\alpha_{ij}^{\Phi_p}$ and $\beta_{\Phi_p}$. Then, the updated node feature $h_i^{(l+1)}$ can be calculated as:
\begin{equation}h_{i}^{\Phi_{p}{ }^{(l)}}=\|_{k=1}^{K} \sigma\left(\textstyle \sum_{j \in \mathcal{N}_{i}^{\Phi p}} \alpha_{i j}^{\Phi_{p}} \cdot \Theta_{\Phi_{j}}^{(l)} \cdot h_{j}^{(l)}\right),\label{eq1}\end{equation}
\begin{equation}h_i^{(l+1)}=\textstyle \sum_{p=1}^P \beta_{\Phi_p}^{(l)} \cdot h_{i}^{\Phi_{p}{ }^{(l)}},\label{eq2}\end{equation}
suppose we denote $\Theta_{\Phi_{j}}^{(l)} \cdot h_{j}^{(l)}$ as ${h_{i}^{'}}^{(l)}$, then $\alpha_{ij}^{\Phi_p}$ and $\beta_{\Phi_p}$ can be computed as:\par
\begin{equation}\alpha_{ij}^{\Phi_p}=softmax_{j}\left(\sigma\left(\theta_{\Phi_p}^{\top} \cdot \left[{h_{i}^{'}}^{(l)}\|{h_{j}^{'}}^{(l)}\right]\right)\right),\label{eq3}\end{equation}
\begin{equation}\beta_{\Phi_p}^{(l)}=softmax_{p}\left(\frac{1}{\left|V\right|}\sum_{v_{i}\in V} \Psi^{\top} tanh\left(W h_{i}^{\Phi_{p}^{(l)}}+b\right)\right).\label{eq4}\end{equation}

\begin{table}
	\caption{Explanations For Notations}
	\label{tab:explanations for notations}
	\centering
	\begin{tabular*}{\hsize}{@{}@{\extracolsep{\fill}}lll@{}}
		\toprule[1.2pt]
		Notations & Explanations \\
		\midrule
		$G_{i}$ & Graph of modality ${i}$ \\
		$V_{i}$ & Set of nodes in modality ${i}$ \\
		$E_{i}$ & Set of edges in modality ${i}$ \\
		$\mathcal{N}_{i}$ & Set of nodes of node type ${i}$ \\
		$N_{i}$ & Number of nodes of node type ${i}$ \\
		$\Phi$ & Meta-path \\
		$X_{i}$ & Feature matrix of $G_{i}$ \\
		$A_{i}$ & Adjacency matrix of $G_{i}$ \\
		${g}$ & Induced graph \\
		${f}$ & Rs-fMRI modality \\
		${d}$ & DTI modality \\
		$\mathbb{R}^{ij}$ & A real vector space of dimension $i \times j$ \\
		${D}$ & Dimension of node features \\
		${\Theta, W, B}$  & Learnable parameter of corresponding layer \\
		\emph{\textbf{O}} & Zero matrix \\
		\bottomrule[1.2pt]
	\end{tabular*}
	\label{tab1}
\end{table}

\section{Materials}
\label{sec:materials}

\subsection{Datasets}
In this study, dataset of MCI is collected from ADNI-3 with 409 subjects, including 100 MCI, 48 early MCI (EMCI), 21 late MCI (LMCI) and 240 normal controls (NC). Every subject included has dual-modal data (i.e., rs-fMRI and DTI). Demographic details of used subjects are shown in Table \ref{tab2}.

\begin{table}
	\caption{Demographic Characteristic of the Used Subjects}
	\label{tab:demographic characteristic of the used subjects}
	\centering
	\begin{tabular*}{\hsize}{@{}@{\extracolsep{\fill}}ccccc@{}}
		\toprule[1.2pt]
		Datasets & \multicolumn{4}{c}{ADNI-3} \\
		\cmidrule{2-5}
		Subject & MCI & EMCI & LMCI & NC \\
		\cmidrule{2-5}
		Number & 100 & 48 & 21 & 240 \\
		Age & 76.73$\pm$8.99 & 75.08$\pm$7.03 & 72.81$\pm$7.14 & 73.76$\pm$7.94 \\
		Gender(M/F) & 49/51 & 28/20 & 8/13 & 96/144 \\
		Manufacturer & SIEMENS & SIEMENS & SIEMENS & SIEMENS \\
		\bottomrule[1.2pt]
	\end{tabular*}
	\label{tab2}
\end{table}

\subsection{Preprocessing}
For rs-fMRI data, the raw rs-fMRI images are preprocessed by Data Processing Analysis for Brain Imaging (DPABI) \cite{b24}. Specifically, 1) the first ten rs-fMRI volumes are discarded as temporal noise. 2) Head movement correction, spatial normalize and smooth filtering are applied furtherly. 3) The Automated Anatomical Labeling (AAL) template is used to segment brain space into 90 brain regions of interest (ROI). 4) Mean time series are extracted for each ROI.\par
For DTI data, the raw DTI images are processed by the pipeline for analyzing brain diffusion images (PANDA) \cite{b25}, respectively. 1) Diffusion metrics are calculated to generate fractional anisotropy (FA) maps. 2) AAL template is used to segment brain space in the FA maps into 90 brain ROIs for further radiomic feature extraction. 3) Deterministic fiber tractography is employed to track white matter fibers. Through the above process, we obtain 90 ROIs with radiomic features \cite{b26} for each subject.

\begin{figure*}[!t]
	\centerline{\includegraphics[width=\textwidth]{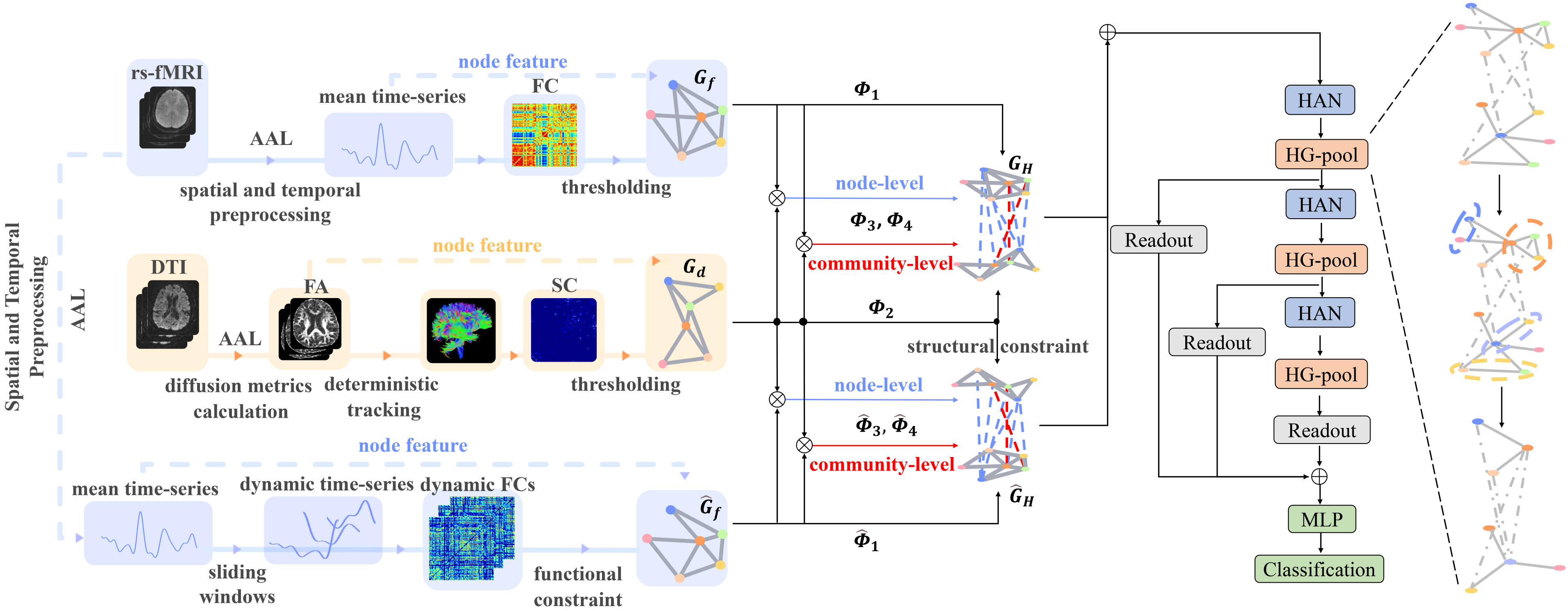}}
	\caption{Overview of our proposed method. a) We extract node features, $\Phi_{1}$ and $\Phi_{2}$ from each modality to establish $G_{f}=\left\{\mathcal{N}_{f}, \Phi_{1}\right\}$, $G_{d}=\left\{\mathcal{N}_{d}, \Phi_{2}\right\}$. b) Node-level and community-level hetero-meta-paths are combined as meta-path $\Phi_{3}:\mathcal{N}_{f} \rightarrow \mathcal{N}_{d}$, and $\Phi_{4}$ is a reversal of $\Phi_{3}$. The subject-level HG is denoted as $G_{H}=\left\{\left(\mathcal{N}_{f}, \mathcal{N}_{d}\right), \left(\Phi_{1}, \Phi_{2}, \Phi_{3}, \Phi_{4}\right)\right\}$,. c) We preserve $\Phi_{2}$ and dynamically reconstruct FC to obtain $\hat{\Phi}_{1}$, then update $\Phi_{3}$ and $\Phi_{4}$ to generate augmented $\hat{G}_{H}$. d) Both $G_{H}$ and $\hat{G}_{H}$ are fed into backbone consisted of HAN, HG pooling and readout layers to extract dual-modal features.}
	\label{fig1}
\end{figure*}

\section{Methods}
\label{sec:methods}
In present work, we propose a novel HGNN to fuse dual-modal information. We define meta-paths in the fused $G_{H}$ as $\Phi_{1}:\mathcal{N}_{f} \rightarrow \mathcal{N}_{f}$, $\Phi_{2}:\mathcal{N}_{d} \rightarrow \mathcal{N}_{d}$, $\Phi_{3}:\mathcal{N}_{f} \rightarrow \mathcal{N}_{d}$, and $\Phi_{4}:\mathcal{N}_{d} \rightarrow \mathcal{N}_{f}$, where homo-meta-paths $\Phi_{1}$, $\Phi_{2}$ are edges of FC or SC, and hetero-meta-paths $\Phi_{3}$, $\Phi_{4}$ are edges between FC and SC. Then the augmented $\hat{G_{H}}$ is generated with reconstructed  $\hat{\Phi_{1}}$ and preserved $\Phi_{2}$. Both $G_{H}$ and $\hat{G_{H}}$  are fed into backbone consisted of HAN, HG pooling and readout layers to extract dual-modal features. Finally, a multilayer perception (MLP) is designed to perform classification with dual-modal features. Detailed overview of proposed framework is demonstrated in Fig. \ref{fig1}.

\begin{figure}[!t]
	\centerline{\includegraphics[width=\columnwidth]{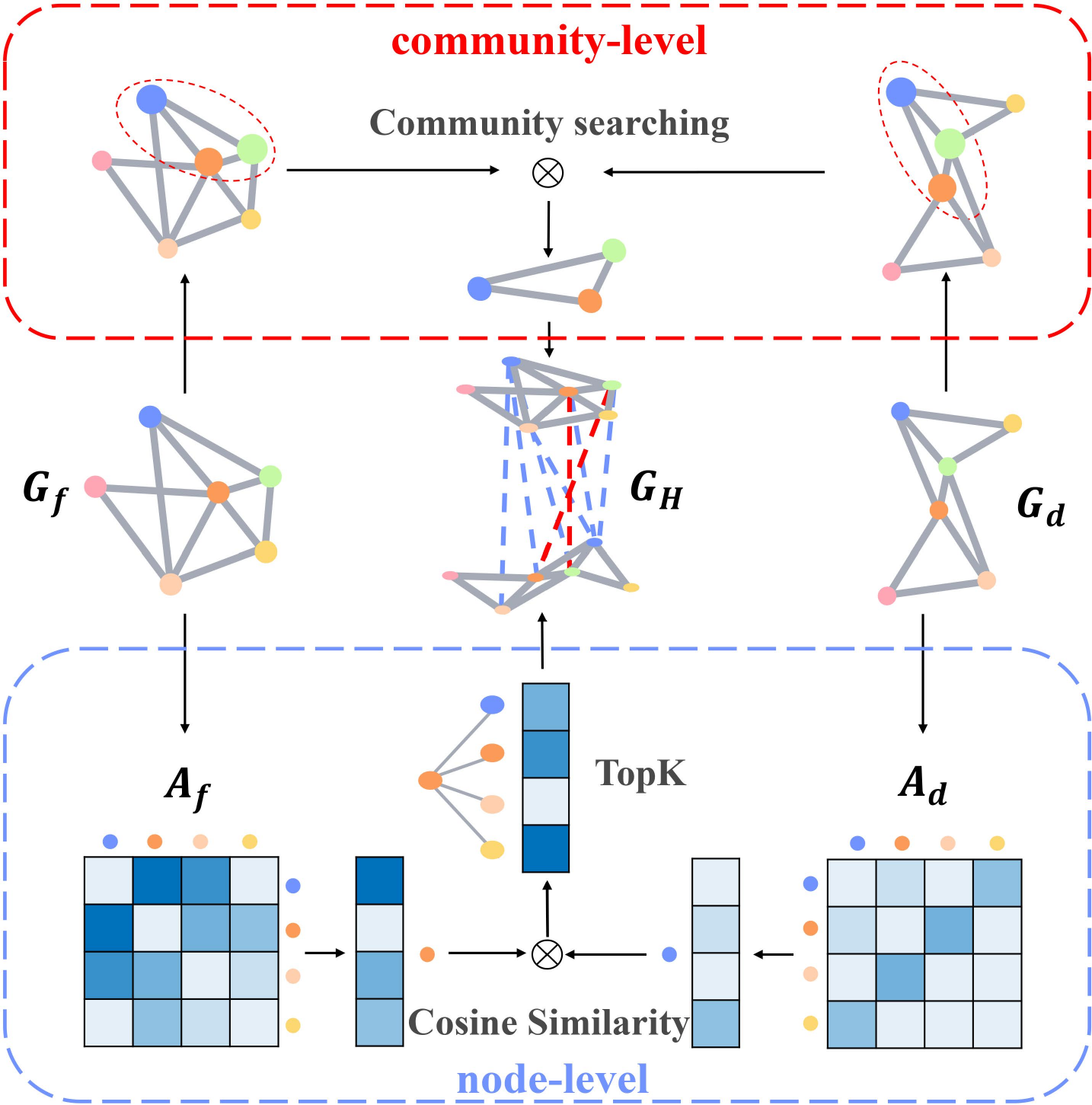}}
	\caption{Overview of node-level and community-level hetero-meta-path.}
	\label{fig2}
\end{figure}

\subsection{Construction of subject-level Heterogeneous Graph}
We propose to construct a subject-level heterogeneous graph $G_{H}=\left\{X_{H}, A_{H}\right\}$ for each subject and perform graph-level classification, and the core works are extracting node features for both modalities and establishing meta-paths. Specifically, we first extract node features and construct edges of $\Phi_{1}$ and $\Phi_{2}$ in each modality, then we construct edges of $\Phi_{3}$ and $\Phi_{4}$ from node-level and community-level. Detailed technics are summarized as follows:

\subsubsection{Construction of homo-meta-path}
For rs-fMRI images, we denote the mean time-series of the ${i}$-th ROI as $TS_{i}^{mean} \in \mathbb{R}^{1 \times D}$. We consider each ROI as a node and $TS_{i}^{mean}$ of corresponding ROI as its node feature ${X_{f}}_{i}$, where ${X_{f}}_{i}$ is the ${i}$-th row vector of $X_{f} \in \mathbb{R}^{N_{f} \times D}$. Subsequently, we calculate the Pearson correlation coefficient values between $TS_{i}^{mean}$ of ROIs to build a FC for each subject. After filtering out edges with attributes lower than 0.2, FC is normalized to obtain $A_{f} \in \mathbb{R}^{N_{f} \times N_{f}}$ corresponding to $\Phi_{1}:\mathcal{N}_{f} \rightarrow \mathcal{N}_{f}$. $G_{f}$ is used to describe both $X_{f}$ and $A_{f}$ for convenience.\par
For DTI images, we treat each ROI as a node with extracted radiomic features \cite{b26} from FA maps as node feature ${X_{d}}_{i}$, where ${X_{d}}_{i}$ is the i-th row vector of $X_{d} \in \mathbb{R}^{N_{d} \times D}$. Subsequently, we track white matter fibers connecting ROIs to build a SC. After filtering out edges with attributes lower than 5, SC is normalized to obtain $A_{d} \in \mathbb{R}^{N_{d} \times N_{d}}$ corresponding to $\Phi_{2}:\mathcal{N}_{d} \rightarrow \mathcal{N}_{d}$. $G_{d}$  is used to describe both $X_{d}$ and $A_{d}$ for convenience.

\subsubsection{Construction of hetero-meta-path}
In constructing node-level hetero-meta-path, we measure the similarity of connection patterns of cross-modal node pairs as the strength of their structure-function coupling. Specifically, for a node $v_{i}$ in $G=\left\{X, A\right\}$, we refer the ${i}$-th row vector $A_{i}$ of ${A}$ to its connection pattern. Then, for a cross-modal node pair $\left({v_{f}}_{i}, {v_{d}}_{j}\right)$, supposing $\varepsilon_{ij}$ denotes the similarity value of their connection patterns, we can obtain a series of $\varepsilon_{ij}$ for node ${v_{f}}_{i}$ denoted as a vector $\vec{\varepsilon}_{i}$. Considered of sparsity, we only construct edges between cross-modal node pairs with most similar connection patterns, which is formulated as:
\begin{equation}{A_{fd}^{(node)}}_{ij}=\left\{\begin{matrix}
		\varepsilon_{ij}, & j \in TopK_{j}\left(\vec{\varepsilon}_{i}\right)\\
		0, & else
	\end{matrix}\right., i=1,2,\dots,N_{f},\label{eq5}\end{equation}
where K equals to 8 in our work to filter out edges with less structure-function coupling strength, $A_{fd}$ corresponds to $\varPhi_{3}$.\par
As for community-level hetero-meta-path, we suggest that brain regions with cooperative interactions may form a closed induced subgraph in both $G_{f}$ and $G_{d}$. Technically, we denote a closed induced subgraph consisted of three nodes as $g_{closed}=\left\{v_{i,j,k},e_{ij,jk,ki}\right\}$. If $g_{closed}$ exists in both $G_{f}$ and $G_{d}$, we add $e_{ij,jk,ki}$ into $A_{fd}$ considering the first node from $G_{f}$ and the second one from $G_{d}$, which is formulated as:
\begin{equation}{A_{fd}^{(community)}}_{ij}=\left\{\begin{matrix}
		1, & g_{closed} \in G_{f} \cap G_{d}\\
		0, & else
	\end{matrix}\right..\label{eq6}\end{equation}\par
Combining the both hetero-meta-paths, we can establish $\varPhi_{3}$ as:
\begin{equation}A_{fd}=Norm\left(A_{fd}^{(node)}+A_{fd}^{(community)}\right).\label{eq7}\end{equation}\par
Combining all the adjacency matrices of meta-paths and node feature matrices, we can build a subject-level heterogeneous graph $G_{H}$, with $X_{H}$, $A_{H}$ calculated as:
\begin{equation}X_{H}=\begin{pmatrix}
		X_{f}\\
		X_{d}
	\end{pmatrix},\label{eq8}\end{equation}
\begin{equation}A_{H}=\begin{pmatrix}
		A_{f} & A_{fd}\\
		A_{fd}^\top & A_{d}
	\end{pmatrix}.\label{eq9}\end{equation}

\subsection{Heterogeneous Graph Augmentation}
The abundant heterogeneity of the HG provides ample possibilities from the perspective of construction, which provides convenience for augmentation. Therefore, we propose to dynamically reconstruct FC to obtain $\hat{\Phi}_{1}$, then $\Phi_{3}$, $\Phi_{4}$ will naturally update along with $\Phi_{1}$. While $\Phi_{2}$ is fixed as structural constraint to maintain the semantic consistency of HGs before and after augmentation.\par
Specifically, in reconstructing FC, we apply the sliding window method to the mean time-series of each ROI to obtain a series of dynamic window time-series with total number $n_{win}$. Then, we calculate the Pearson correlation coefficient values among the dynamic window time-series of the ${w}$-th window to build a local dynamic FC, denoted as $FC_{L}^{w}$.\par
To ensure the consistency in HG construction method, we need to summarize the entire series of $FC_{L}^{w}$ into one global dynamic FC, denoted as $FC_{G}$. To preserve key dynamic features from all the $FC_{L}^{w}$, we search shared induced subgraph ${g}$ consisted of three nodes which contains edge $e_{ij}$ and exists in each $FC_{L}^{w}$, and summarize it into $FC_{G}$. However, the topology of ${g}$ varies based on total number of edges within it, which can be denoted as $g\left(n\right)$, where ${n}$ is the number of edges. Apparently, the importance of $g\left(n\right)$ varies along with its topology, which directly influences the richness of feature interaction within it. Therefore, we set different attention weights for different $g\left(n\right)$. Then, we can calculate the attributes of $e_{ij}$ in $FC_{G}$, which is formulated as:
\begin{equation}{FC_{G}}_{ij}=\vec{\alpha}^\top \cdot \vec{f}_{g\left(n\right)},\label{eq10}\end{equation}
where $\vec{a} \in \mathbb{R}^{3\times1}$ is the attention weight vector, and $\vec{f}_{g\left(n\right)}\in\mathbb{R}^{3\times1}$ is the vector of frequency of occurrence of different $g\left(n\right)$. We set $\vec{\alpha}^\top=\left[0.01,0.02,0.1\right]$ in present work, which is inversely proportional to the frequency of occurrence of corresponding $g\left(n\right)$. Subsequently, we filter out edges with attributes lower than 0.4 to ensure the sparsity of $FC_{G}$.\par
Finally, we consider edges in $FC_{G}$ as $\hat{\Phi}_{1}$ corresponding to $\hat{A}_{f} \in \mathbb{R}^{N_{f} \times N_{f}}$. With $\Phi_{2}$ fixed, we can update $\Phi_{3}$ and $\Phi_{4}$ following \eqref{eq5}-\eqref{eq7}. Then the augmented $\hat{G}_{H}$ can be constructed following \eqref{eq8}-\eqref{eq9}. We sent $G_{H}$ and $\hat{G}_{H}$ in pair into the backbone to avoid data leakage.

\begin{figure}[!t]
	\centerline{\includegraphics[width=\columnwidth]{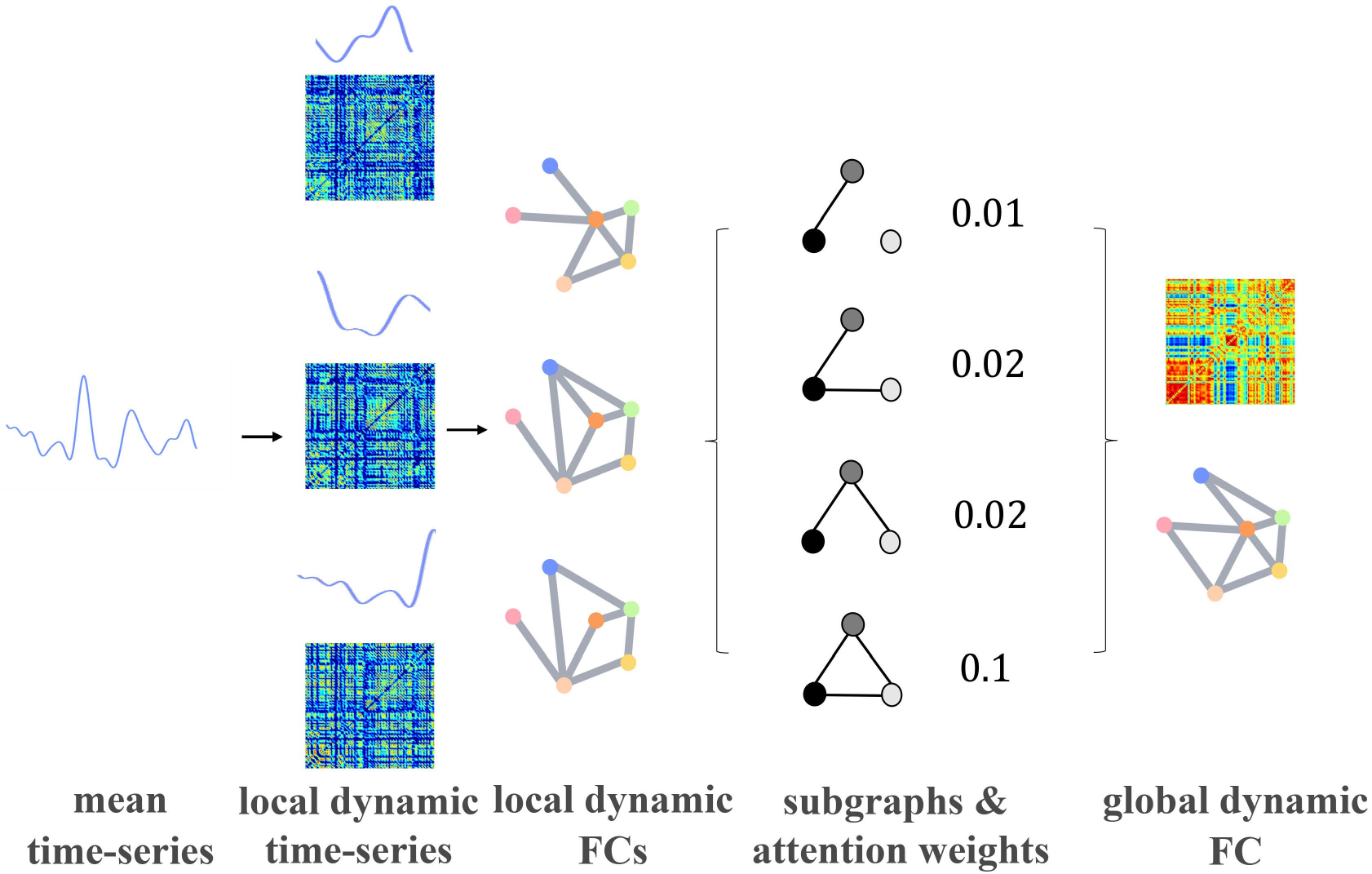}}
	\caption{Overview of dynamically reconstructed FC in HG augmentation.}
	\label{fig3}
\end{figure}

\begin{figure}[!t]
	\centerline{\includegraphics[width=\columnwidth]{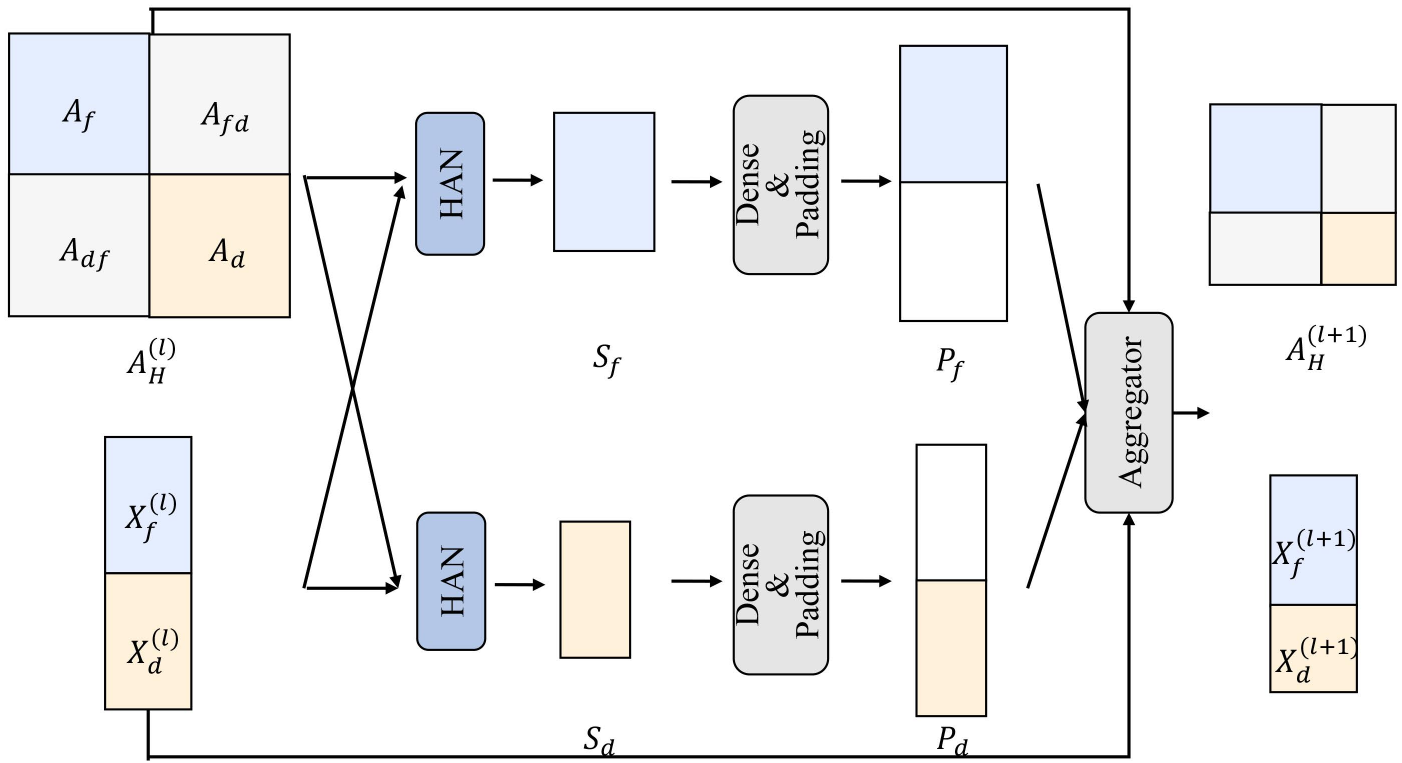}}
	\caption{Overview of introduced HG pooling strategy.}
	\label{fig4}
\end{figure}

\subsection{Heterogeneous Graph Pooling}
We introduce a heterogeneous graph pooling strategy \cite{b27} into our framework to better leverage heterogeneous information and avoid feature confusion. Specifically, we calculate initial pooling score matrices considering both heterogeneous topology and node features. Then, we summarize the initial pooling matrices by a dense layer to avoid feature confusion among different kinds of nodes.\par
Technically, taking a subject-level HG with initial adjacency matrix $A_{H}^{(0)} \in \mathbb{R}^{N \times N}$ and feature matrix $X_{H}^{(0)} \in \mathbb{R}^{N \times D}$ for example, where $N=N_{f}+N_{d}$ is the total number of nodes. We denote the output by the ${l}$-th layer as $A_{H}^{(l)} \in \mathbb{R}^{N^{(l)} \times N^{(l)}}$ and $X_{H}^{(l)} \in \mathbb{R}^{N^{(l)} \times D}$. Different from the original method, we propose to apply a HAN layer to $A_{H}^{(l)}$ and $X_{H}^{(l)}$ to learn pooling score matrices for different types of nodes separately, which not only considers heterogeneous node features and topology but also can apply different pooling ratios to different types of nodes. Taking the pooling for nodes of rs-fMRI modality for instance, we calculate the initial pooling score matrix $S_{f}^{(l)}$ as:
\begin{equation}S_{f}^{(l)}=HAN\left(A_{H}^{(l)}, X_{H}^{(l)}; \Theta_{f}^{(l)}\right).\label{eq11}\end{equation}\par
However, directly applying $S_{f}^{(l)}$ to $A_{H}^{(l)}$ and  $X_{H}^{(l)}$ will lead to feature confusion. Therefore, we need to summarize $S_{f}^{(l)}$ into $D_{f}^{(l)}$ with node-specific pooling information as:
\begin{equation}D_{f}^{(l)}=softmax\left(W_{f}^{(l)}S_{f}^{(l)}+B_{i}^{(l)}\right),\label{eq12}\end{equation}
where $W_{f}^{(l)} \in \mathbb{R}^{N_{f}^{(l+1)} \times N^{(l)}}$ and $B_{i}^{(l)} \in \mathbb{R}^{N_{f}^{(l+1)} \times N_{f}^{(l+1)}}$.To avoid extra dimensional operations, we apply zero paddings to the first dimension of $D_{f}^{(l)}$ to obtain the final pooling score matrix $P_{f}^{(l)}$, which is formulated as:
\begin{equation}P_{f}^{(l)}=\begin{pmatrix}
		D_{f}^{(l)}\\
		\emph{\textbf{O}}
	\end{pmatrix},\label{eq13}\end{equation}
where $\emph{\textbf{O}} \in \mathbb{R}^{N_{d}^{(l+1)} \times N_{f}^{(l+1)}}$ and $P_{f}^{(l)} \in \mathbb{R}^{N^{(l)} \times N_{f}^{(l+1)}}$.\par
Finally, we apply $P_{f}^{(l)}$, $P_{d}^{(l)}$ to $A_{H}^{(l)}$ and $X_{H}^{(l)}$ to accomplish HG pooling:
\begin{equation}A_{H}^{(l+1)}=\begin{pmatrix}
		{P_{f}^{(l)}}^\top A_{f}^{(l)} P_{f}^{(l)} & {P_{f}^{(l)}}^\top A_{fd}^{(l)} P_{d}^{(l)}\\
		{P_{d}^{(l)}}^\top {A_{fd}^{(l)}}^\top P_{f}^{(l)} & {P_{d}^{(l)}}^\top A_{d}^{(l)} P_{d}^{(l)}
	\end{pmatrix},\label{eq14}\end{equation}
\begin{equation}X_{H}^{(l+1)}=\begin{pmatrix}
		{P_{f}^{(l)}}^\top X_{f}^{(l)}\\
		{P_{d}^{(l)}}^\top X_{d}^{(l)}
	\end{pmatrix},\label{eq15}\end{equation}
where $A_{H}^{(l+1)} \in \mathbb{R}^{N^{(l+1)} \times N^{(l+1)}}$ and $X_{H}^{(l+1)} \in \mathbb{R}^{N^{(l+1)} \times D}$.

\begin{table*}
	\caption{Results of Performance Evaluation Experiments}
	\label{tab:evaluation experiments}
	\centering
	\begin{tabular*}{\hsize}{@{}@{\extracolsep{\fill}}ccllllcccc@{}}
		\toprule[1.2pt]
		Ref & Year & Modality & Subject & Method & Task & ACC(\%) & SEN(\%) & SPE(\%) & AUC \\
		\midrule[1pt]
		\cite{b30} & 2021 & fMRI & 191MCI, 179NC & MMTGCN & NC vs. MCI & 86.0 & 86.9 & 85.1 & 0.903 \\
		\midrule
		\cite{b31} & 2022 & fMRI & 68MCI, 69NC & BFN & NC vs. MCI & 90.5 & 91.2 & 89.9 & 0.952 \\
		\midrule
		\multirow{2}*{\cite{b32}} & \multirow{2}*{2022} & \multirow{2}*{fMRI} & \multirow{2}*{72EMCI, 53LMCI, 64NC} & \multirow{2}*{dEC+cwGAT} & NC vs. EMCI & 90.9 & 90.4 & 91.4 & 0.967 \\
		~ & ~ & ~ & ~ & ~ & EMCI vs. LMCI & 89.8 & 87.6 & 91.4 & 0.940 \\
		\midrule
		\multirow{2}*{\cite{b33}} & \multirow{2}*{2024} & \multirow{2}*{fMRI} & \multirow{2}*{168EMCI, 120LMCI, 154NC} & \multirow{2}*{MGCA-RAFFNet} & NC vs. EMCI & 92.2 & 92.2 & 92.3 & 0.930 \\
		~ & ~ & ~ & ~ & ~ & EMCI vs. LMCI & 91.3 & 91.7 & 90.8 & 0.920 \\
		\midrule
		\cite{b34} & 2020 & fMRI+DTI & 36MCI, 37NC & Adaptive dFC & NC vs. MCI & 87.7 & 88.9 & 86.5 & 0.889 \\
		\midrule
		\cite{b35} & 2024 & fMRI+DTI & 151MCI, 142NC & Cross-GNN & NC vs. MCI & 82.6 & 84.6 & 82.2 & / \\
		\midrule
		\multirow{3}*{\cite{b36}} & \multirow{3}*{2021} & \multirow{3}*{fMRI+DTI} & \multirow{3}*{44EMCI, 38LMCI, 44NC} & \multirow{3}*{SAC-GCN} & NC vs. EMCI & 85.2 & 90.9 & 79.5 & 0.898 \\
		~ & ~ & ~ & ~ & ~ & NC vs. LMCI & 89.0 & 89.5 & 88.6 & 0.928 \\
		~ & ~ & ~ & ~ & ~ & EMCI vs. LMCI & 86.6 & 92.1 & 81.8 & 0.943 \\
		\midrule
		\multirow{3}*{\cite{b9}} & \multirow{3}*{2023} & \multirow{3}*{fMRI+DTI} & \multirow{3}*{86EMCI, 166LMCI, 163NC} & \multirow{3}*{MMP-GCN} & NC vs. EMCI & 91.2 & 82.6 & 95.7 & / \\
		~ & ~ & ~ & ~ & ~ & NC vs. LMCI & 94.2 & 92.8 & 95.7 & / \\
		~ & ~ & ~ & ~ & ~ & EMCI vs. LMCI & 92.4 & 93.7 & 89.5 & / \\
		\midrule
		\multirow{4}*{Ours} & \multirow{4}*{2024} & \multirow{4}*{fMRI+DTI} & \multirow{4}*{100MCI, 48EMCI, 21LMCI, 240NC} & \multirow{4}*{Brain-HAN} & NC vs. MCI & \textbf{93.3} & \textbf{93.5} & \textbf{93.3} & \textbf{0.961} \\
		~ & ~ & ~ & ~ & ~ & NC vs. EMCI & \textbf{92.7} & \textbf{90.6} & \textbf{96.9} & \textbf{0.927} \\
		~ & ~ & ~ & ~ & ~ & NC vs. LMCI & \textbf{94.4} & \textbf{92.1} & \textbf{97.0} & \textbf{0.972} \\
		~ & ~ & ~ & ~ & ~ & EMCI vs. LMCI & \textbf{92.4} & \textbf{85.7} & \textbf{96.0} & \textbf{0.952} \\
		\bottomrule[1.2pt]
	\end{tabular*}
	\label{tab3}
\end{table*}

\begin{figure*}[!t]
	\centerline{\includegraphics[width=\textwidth]{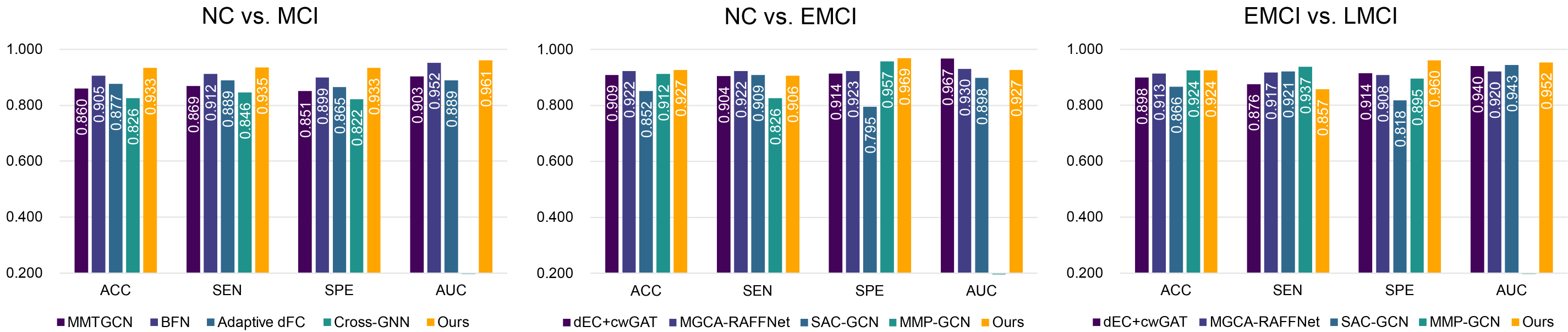}}
	\caption{Bar-chart of performance evaluation with SOTA methods.}
	\label{fig5}
\end{figure*}

\section{Experiments And Results}
\label{sec:experiments and results}
We conduct performance evaluation experiments with state-of-the-art related works on the performance on MCI identification and ablation experiments to verify the effect of proposed mechanisms. In our experiments, backbone consists of three HAN layers, three HG pooling layers and three readout layers. PairNorm \cite{b28} strategy is introduced to tackling the oversmoothing during training. Readout layer concatenates the max and mean values of node features to obtain graph embeddings. We concatenate graph embeddings of three readout layers as the input for MLP to perform classification. Adam optimizer is used to optimize our network with initial learning rate $1 \times 10^{-4}$ and initial weight decay $1 \times 10^{-4}$. A cosine annealing strategy \cite{b29} is introduced during training. We set the number of attention heads in HAN layers to 8, the dimension of hidden layers to 128, the pooling ratio of pooling layers to 0.8, the dropout rate to 0.45, the number of total epochs to 200, and the batch size to 32. Based on the 5-fold cross-validation strategy, prediction accuracy (ACC), sensitivity (SEN), specificity (SPE) and area under the curve (AUC) are calculated as evaluation metrics.

\begin{table*}
	\caption{Results of Ablation Experiments}
	\label{tab:ablation experiments}
	\centering
	\begin{tabular*}{\hsize}{@{}@{\extracolsep{\fill}}lcccccccc@{}}
		\toprule[1.2pt]
		\multirow{2}*{Method} & \multicolumn{4}{c}{NC vs. MCI} & \multicolumn{4}{c}{NC vs. EMCI} \\
		\cmidrule{2-9}
		~ & ACC & SEN & SPE & AUC & ACC & SEN & SPE & AUC \\
		\midrule
		baseline & 0.80 & 0.46 & 0.94 & 0.66 & 0.74 & 0.65 & 0.77 & 0.78 \\
		baseline+pool & 0.86 & 0.66 & 0.98 & 0.76 & 0.85 & 0.78 & 0.91 & 0.85 \\
		baseline+pool+commu & 0.90 & 0.77 & 0.98 & 0.84 & 0.89 & 0.87 & 0.92 & 0.91 \\
		baseline+pool+commu+aug & \textbf{0.93} & \textbf{0.93} & \textbf{0.93} & \textbf{0.96} & \textbf{0.93} & \textbf{0.91} & \textbf{0.97} & \textbf{0.93} \\
		\midrule[1pt]
		\multirow{2}*{Method} & \multicolumn{4}{c}{NC vs. LMCI} & \multicolumn{4}{c}{EMCI vs. LMCI} \\
		\cmidrule{2-9}
		~ & ACC & SEN & SPE & AUC & ACC & SEN & SPE & AUC \\
		\midrule
		baseline & 0.72 & 0.60 & 0.83 & 0.80 & 0.70 & 0.63 & 0.77 & 0.78 \\
		baseline+pool & 0.82 & 0.70 & 0.90 & 0.86 & 0.77 & 0.67 & 0.86 & 0.80 \\
		baseline+pool+commu & 0.90 & 0.88 & 0.93 & 0.94 & 0.85 & 0.80 & 0.90 & 0.88 \\
		baseline+pool+commu+aug & \textbf{0.94} & \textbf{0.92} & \textbf{0.97} & \textbf{0.97} & \textbf{0.92} & \textbf{0.86} & \textbf{0.96} & \textbf{0.95} \\
		\bottomrule[1.2pt]
	\end{tabular*}
	\label{tab4}
\end{table*}

\subsection{Performance Evaluation}
In Table \ref{tab3}, we compare our method with eight related state-of-the-art (SOTA) methods published after 2021 on the classification performance of MCI, which include:\par
\textbf{MMTGCN} \cite{b30}, \textbf{BFN} \cite{b31}, \textbf{dEC+cwGAT} \cite{b32}, \textbf{MGCA-RAFFNet} \cite{b33}. These methods are among SOTA methods using single rs-fMRI modality. BFN and dEC+cwGAT propose to estimate more effective brain functional network to learn more efficient features from rs-fMRI. While MMTGCN and MGCA-RAFFNet attempt to apply multiple brain parcellation templates to obtain different ROIs to construct various brain FC, which can reduce the impact of selection of brain parcellation template on GNN as well as explore the influence of different brain parcellation templates on brain disorder diagnosis.\par
\textbf{Adaptive dFC} \cite{b34}, \textbf{Cross-GNN} \cite{b35}, \textbf{SAC-GCN} \cite{b36}, \textbf{MMP-GCN} \cite{b9}. These methods are among SOTA methods leveraging dual-modal information. Adaptive dFC follows the connectivity-level fusion way, assigning different penalty parameters to FC based on DTI tractography to estimate more effective dynamic FC. MMP-GCN also proposes constructing FC with a DTI strength penalty term to fuse dual-modal information on connectivity-level. Cross-GNN and SAC-GCN follow feature-level fusion way, where effective mechanisms are proposed to encode feature embeddings in each modality.\par
Although SOTA methods using single fMRI modality have achieved good performance, the ACC is still limited because of insufficient features learning with single modality. Table III shows their ACC ranges from 82.0\% to 92.2\%. We notice that various dual-modal fusion methods are also studied, and their experimental results show that use dual-modal data can improve the classification performance of MCI. Table III shows their ACC ranges from 82.6\% to 93.0\%. However, compared to our method, the ignorance of the heterogeneity of dual-modal data still limits their performance.\par
Besides, in related works we notice that the imbalance of datasets may lead to imbalanced SEN and SPE resulted in significant negative impact on the final classification performance. Hence, we propose the HG augmentation method to address this problem. As demonstrated in Table III, with HG augmentation method involved, we can achieve more balanced SEN and SPE than related works, which improves the overall classification performance, indicating that our augmentation method can generate new data with trustworthy distributions and effectively boost classification performance.

\begin{figure}[!t]
	\centerline{\includegraphics[width=\columnwidth]{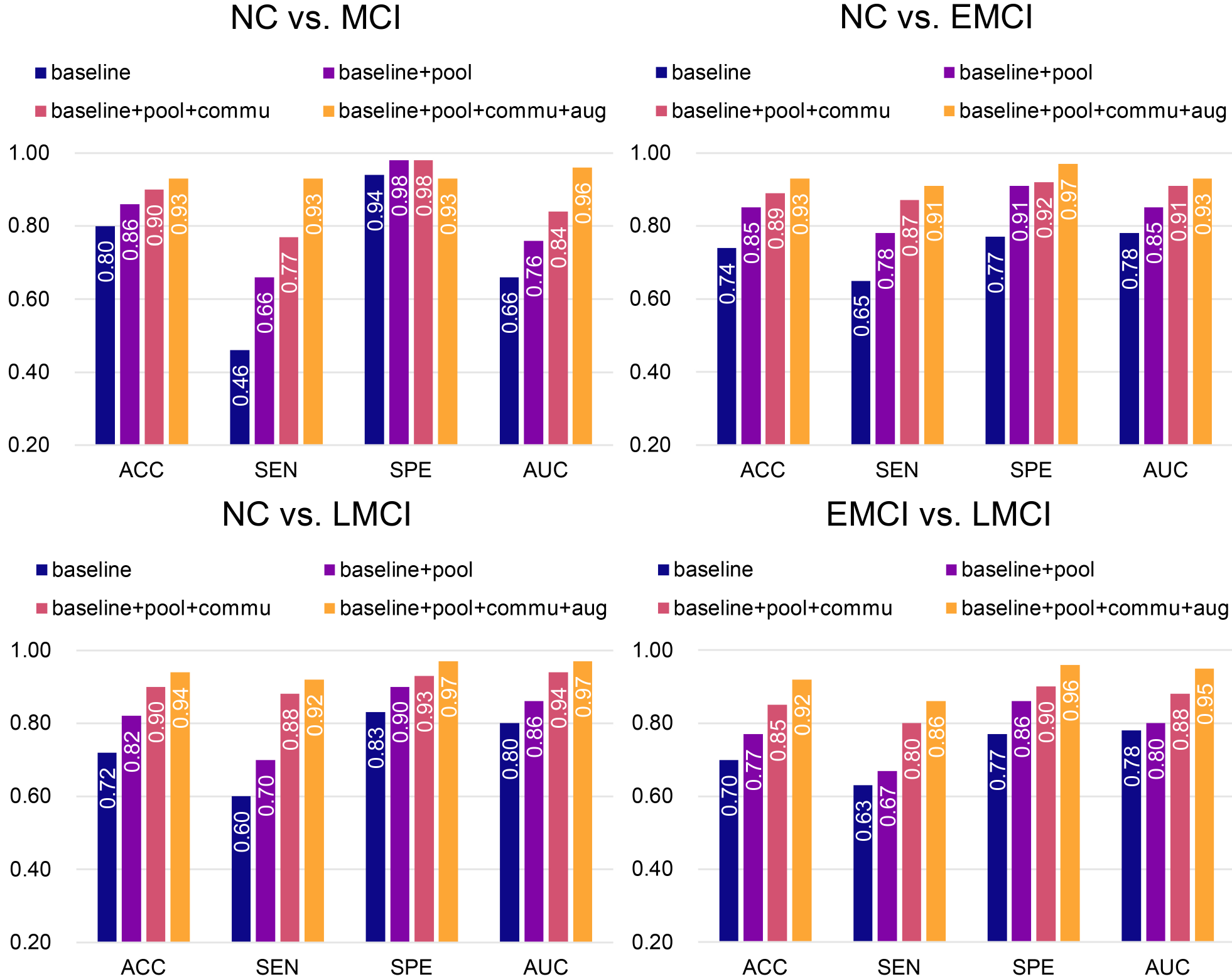}}
	\caption{Bar-chart of ablation experiments.}
	\label{fig6}
\end{figure}

\begin{figure}[!t]
	\centerline{\includegraphics[width=\columnwidth]{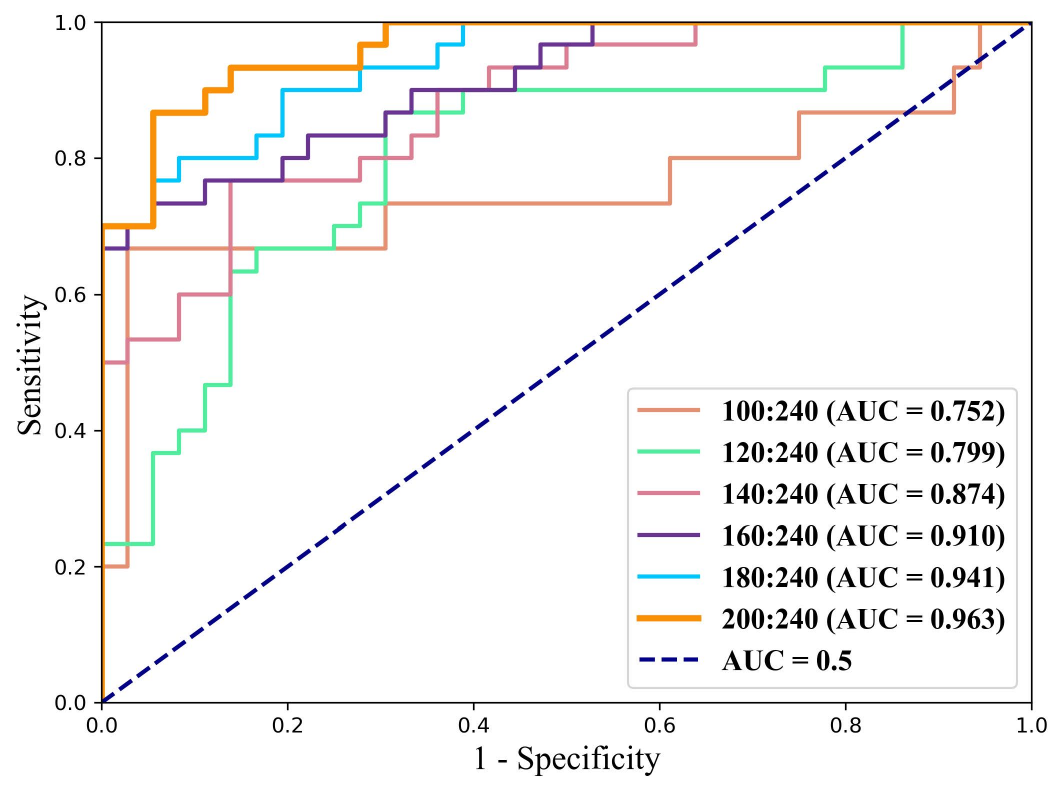}}
	\caption{ROC curves on test dataset with different augmentation ratios.}
	\label{fig7}
\end{figure}

\subsection{Ablation Study}
We further verify the effectiveness of proposed key mechanisms through ablation study in Table \ref{tab4}. The competing methods in the ablation study are listed as follows:\par
1) \textbf{baseline}: $\Phi_{3}$ and $\Phi_{4}$ consist of only node-level hetero-meta-path, and only HAN layers are used to extract features.\par
2) \textbf{baseline + HG pooling} (baseline+pool): This method further involves HG pooling strategy into the backbone.\par
3) \textbf{baseline + HG pooling + community-level hetero-meta-path} (baseline+pool+commu): $\Phi_{3}$ and $\Phi_{4}$ consist of both node-level and community-level hetero-meta-path in this method, and HG pooling strategy is involved in the backbone.\par
4) \textbf{baseline + HG pooling + community-level hetero-meta-path + HG augmentation} (baseline+pool+commu+ aug): HG augmentation method is further involved in this method to balance dataset.\par
First, we verify the effectiveness of the HG pooling strategy. With additional pooling layers, mean ACC improves 11.5\% compared to the baseline and mean SEN has a considerable improvement of 20.1\% in four classification tasks. Next, we verify the effect of the community-level hetero-meta-path. Results of the ablation study show that the community-level hetero-meta-path can effectively improve mean ACC and SEN of 7.3\% and 18.1\% in four classification tasks.\par
However, it’s significant to notice that pooling strategy and community-level hetero-meta-path cannot balance SEN and SPE if the original dataset itself is imbalanced. Take the NC vs. MCI task for instance, dataset of this task suffers from sample imbalance issue with positive and negative sample ratio of 1:2.4. Therefore, we enhance MCI subjects from 100 cases to 200 cases to balance enhanced datasets. According to the results of ablation study, SEN and SPE in NC vs. MCI task both reach over 90\% after augmentation, which brings great improvement of ACC and AUC as well. Table IV shows that mean SEN and SPE improve 9.0\% and 2.7\% in four classification tasks with HG augmentation strategy involved, which indicates that our augmentation method can generate new data with trustworthy distributions. To further discover the influence of different augmentation ratios, we draw the ROC curves on test dataset with framework training with different augmentation ratios in NC vs. MCI task, as shown in Fig. \ref{fig7}. We can notice that AUC of framework improves along with the increase of augmentation ratio.\par
\begin{figure}[!t]
	\centerline{\includegraphics[width=\columnwidth]{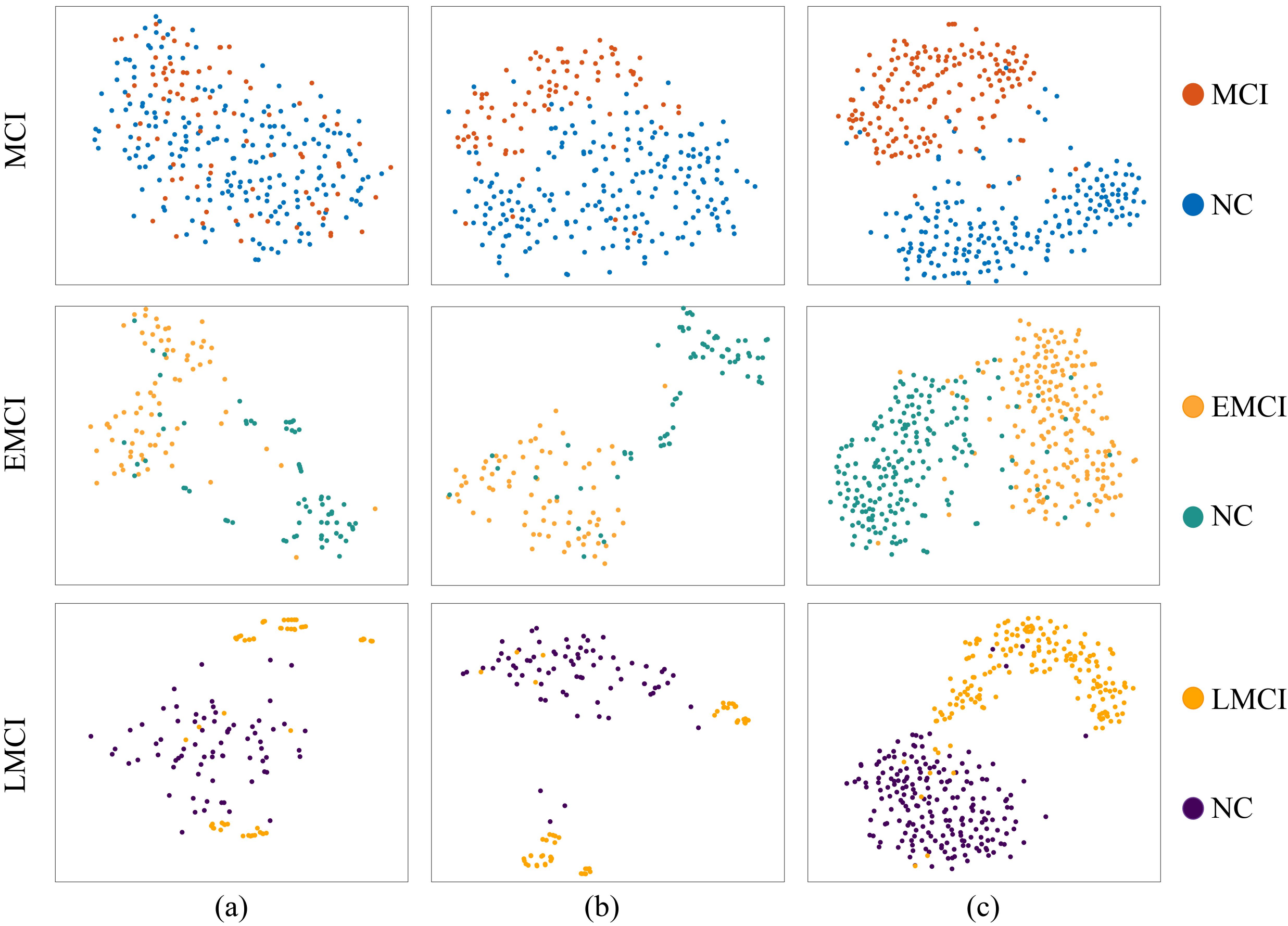}}
	\caption{The t-SNE visualization using different mechanisms. (a) baseline, (b) baseline+pool+commu, (c) baseline+pool+commu+aug.}
	\label{fig8}
\end{figure}

Besides, we perform t-SNE visualization in ablation study. Specifically, we extract the output features of backbone and bring them down to two dimensions to visualize the distribution of learned features. Fig. \ref{fig8} shows that with more mechanisms involved, the distribution of features of different classes is more distinguishable.

\section{Discussion}
\label{sec:discussion}

\subsection{Analysis of Meta-paths}
Previous studies on structure-function coupling have pointed out that structure-function relationship may itself be gradually decoupling from unimodal to transmodal cortex \cite{b37,b38}. This phenomenon can be a great challenge for connectivity-level dual-modal fusion methods based on GNNs, since their fused connectivity can be significantly affected by structure-function decoupling. Therefore, we propose to analyze the meta-paths of HG constructed in present work to find out the impact of them on the phenomenon of structure-function decoupling in the transmodal cortex.\par
\begin{figure}[!t]
	\centerline{\includegraphics[width=\columnwidth]{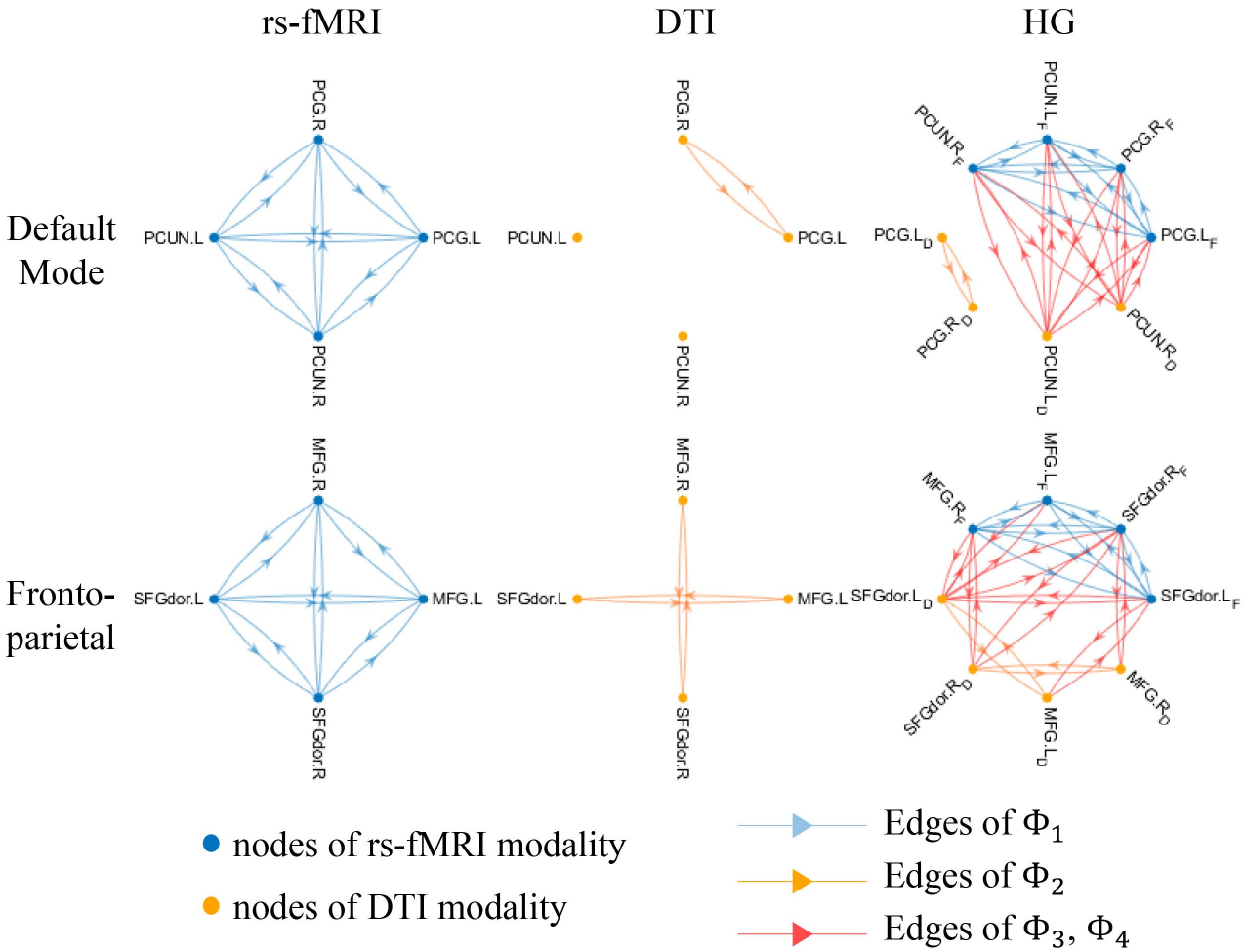}}
	\caption{Visualization of meta-paths within transmodal cortex.}
	\label{fig9}
\end{figure}

In Fig. \ref{fig9}, we visualize meta-paths of our HGs within transmodal cortex consisted of the default mode and fronto-parietal functional networks \cite{b39}. We can notice that with only homo-meta-path, the connection patterns of edges in rs-fMRI and DTI modalities are quite different, which indicates that brain structure and function relations are decoupling in transmodal cortex. While with hetero-meta-path $\Phi_{3}$ and $\Phi_{4}$ involved, brain regions that are particularly separated in DTI modality can be linked together through cross-modal edges, which indicates that our HG-based method can alleviate structure-function decoupling within transmodal cortex.

\begin{figure}[!t]
	\centerline{\includegraphics[width=\columnwidth]{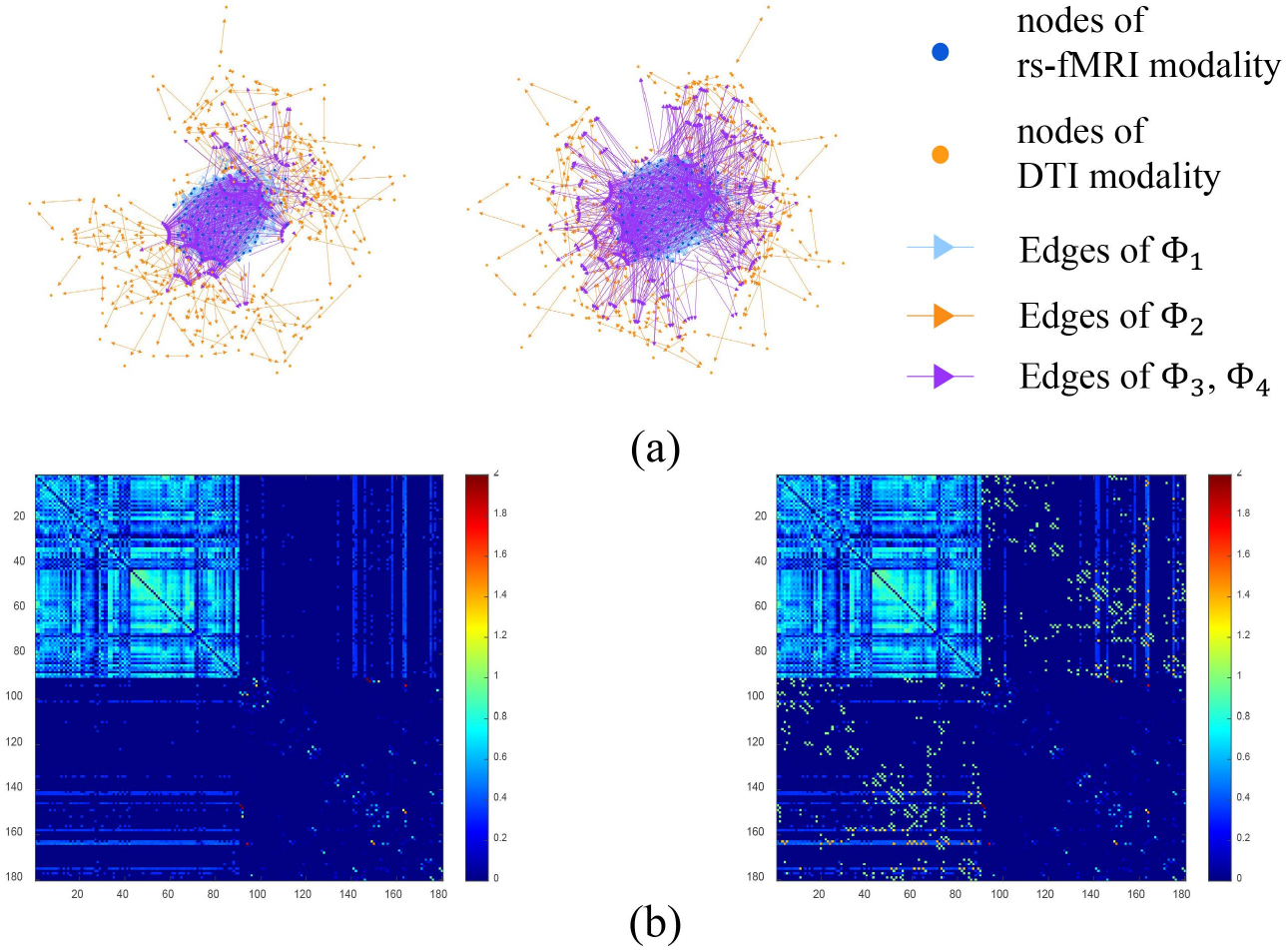}}
	\caption{Comparison of node-level and community-level hetero-meta-path. (a) Graph topologies, (b) adjacency matrices.}
	\label{fig10}
\end{figure}

\subsection{Analysis of Node-level and Community-level Hetero-meta-path}
In Fig. \ref{fig10}, we visualize the topology of HG constructed with only node-level hetero-meta-path, HG constructed with both hetero-meta-paths, and their corresponding adjacency matrices.\par
In Fig. \ref{fig10}a, we can notice that edges of $\Phi_{3}$ and $\Phi_{4}$ significantly increase when community-level hetero-meta-path is involved, which greatly enriches the heterogeneity of HG. Besides, we notice that when there is only node-level hetero-meta-path, edges of $\Phi_{3}$ will only connect with nodes with high degrees in DTI modality. While with community-level hetero-meta-path established, edges of $\Phi_{3}$ will spread to nodes with less degrees. This phenomenon is more obvious in Fig. \ref{fig10}b, where only striped connectivity exists in $\Phi_{3}$ and $\Phi_{4}$ with only node-level hetero-meta-path, while the involvement of community-level hetero-meta-path creates more centrally symmetric scatter connectivity. Considering of structure-function decoupling, it is nature that node-level hetero-meta-path can only capture relations within brain regions with strong structure-function coupling. On the other hand, community-level hetero-meta-path concentrated on shared brain communities can avoid being disturbed by structure-function decoupling, which can increase the diversity of heterogeneous connectivity. Hence, combining both node-level and community-level hetero-meta-path captures more sufficient dual-modal information to ensure the robustness of HG.

\begin{figure}[!t]
	\centerline{\includegraphics[width=\columnwidth]{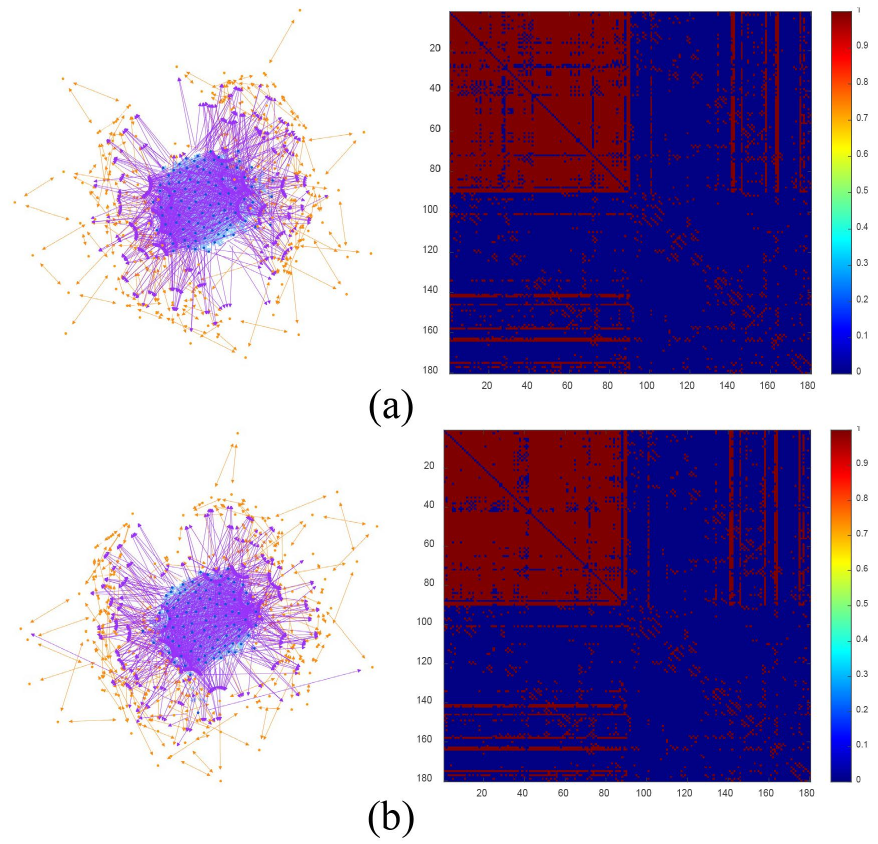}}
	\caption{Visualization of HGs before and after augmentation. (a) original $G_{H}$ and $A_{H}$ (b) augmented $\hat{G}_{H}$ and $\hat{A}_{H}$.}
	\label{fig11}
\end{figure}

\begin{figure}[!t]
	\centerline{\includegraphics[width=\columnwidth]{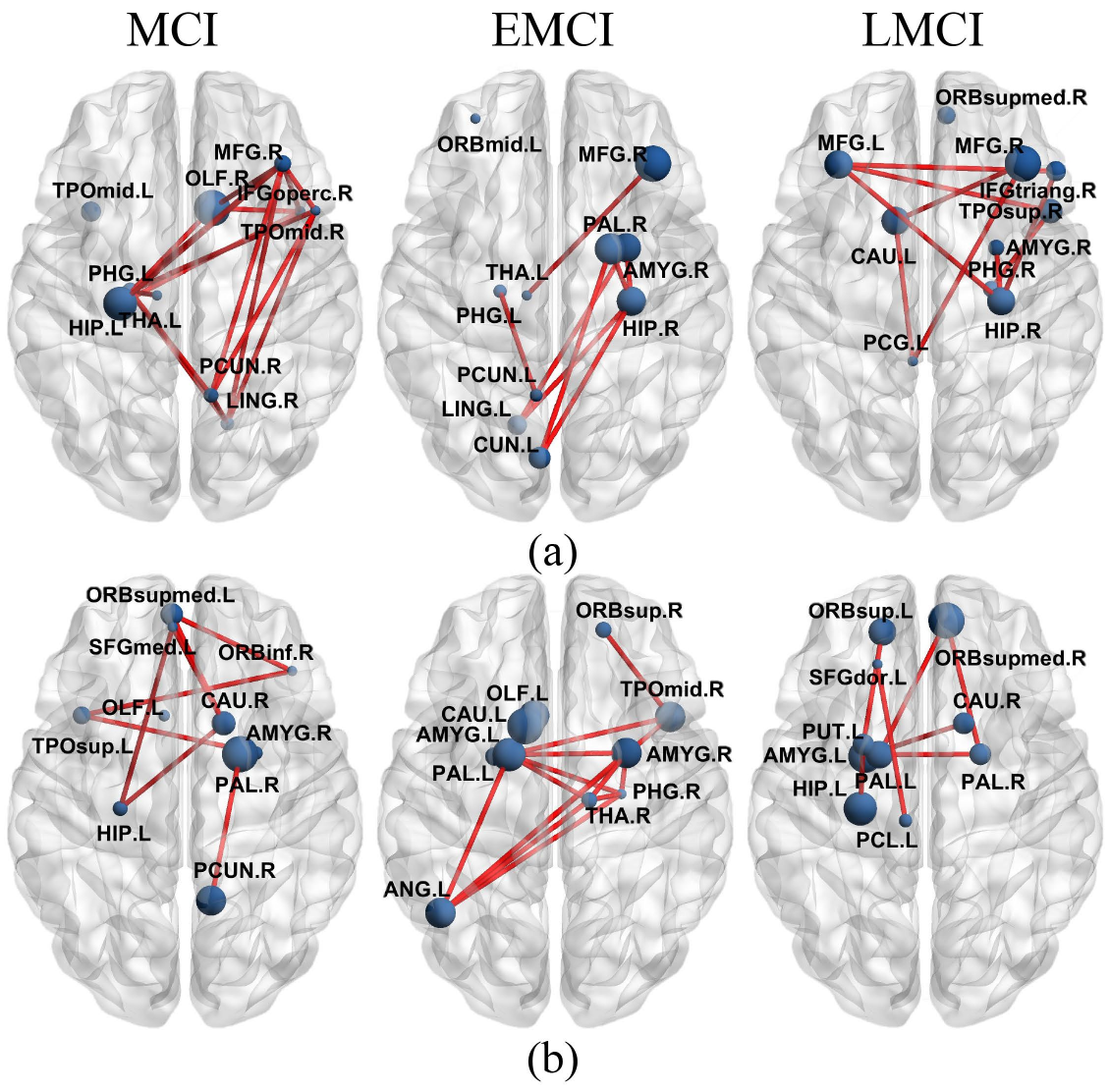}}
	\caption{Visualization of ROIs selected by pooling layer. (a) Selected ROIs in rs-fMRI modality, (b) Selected ROIs in DTI modality.}
	\label{fig12}
\end{figure}

\subsection{Analysis of Heterogeneous Graph Augmentation}
In Fig. \ref{fig11}, We visualize the topologies of $G_{H}$, $\hat{G}_{H}$, $A_{H}$, and $\hat{A}_{H}$. From the comparison of topology of $G_{H}$ and $\hat{G}_{H}$, we can notice that while there are differences in details, the primary topology of HG still preserves after augmentation. The comparison of $A_{H}$ and $\hat{A}_{H}$ provides another intuitive perspective of the change in data distribution. In the upper-left corner of $A_{H}$, we notice important topological features are retained with undergoing changes in details. On the other hand, thanks to the retention of $\Phi_{2}$, updated $\Phi_{3}$ and $\Phi_{4}$ still preserve important topological features as shown in $\hat{A}_{H}$.\par
Above analysis illustrates that proposed HG data augmentation method can generate new HGs with trustworthy distributions and retain important topological features.

\subsection{Most Discriminative ROIs and Connectivity}
Considering that the first pooling layer is directly associated with actual brain regions, we analyze the output of the first pooling layer to identify the important brain regions and connectivity on which the modal relies during inference. The visualization results are shown in Fig. \ref{fig12}.\par
Based on the output of the first pooling layer, we select ROIs that are not clustered into same class as most discriminative regions in rs-fMRI and DTI modalities and use the clustered classes after pooling layer to describe their connectivity. We notice that some ROIs are important in each classification task. For instance, the middle frontal gyrus (MFG) is selected in rs-fMRI modality for every classification task. Previous studies indicate that MFG is one key region of the default mode network invovled many neurophysiological processes of cognitive and emotional function \cite{b40}. Other significant brain regions which has been demonstrated to be assosicated with cognitive like superior frontal gyrus (SFG) and middle temporal gyrus (MTG) \cite{b33,b40} are also selected by our model. Besides, regions from limbic system like amygdala (AMYG) and hippocampus (HIP) are selected as well.\par
On the other hand, we notice that more brain regions from basal ganglia (BG) are selected in DTI than rs-fMRI modality, such as caudate nucleus (CAU) and lenticular nucleus, pallidum (PAL). Previous studies stress that white matter features provided by DTI images can help locate structural changes of BG \cite{b41}, while fMRI are based on cortical signal resulted in more emphasis on the cortex rather than BG. This difference in selected brain regions in our work affirms the conclusion of previous studies.

\section{Conclusion}
\label{sec:conclusion}
In this paper, we propose a HG-based dual-modal fusion method. We define homo- and hetero-meta-path based on domain knowledge to capture important relations, and establish hetero-meta-path from node-level and community-level. We propose HG pooling strategy to better leverage heterogeneous information and avoid feature confusion. A HG augmentation strategy is proposed to address the problem of sample imbalance. We conduct a series of experiments on ANDI-3 dataset, and the results indicate that our method is superior to existing SOTA methods. Furthermore, ablation experiments and discussion show that proposed method has promising generalization ability and clinical interpretability. In the future, we will further discover HG construction method and apply proposed method to other types of brain disorders.

\end{document}